\documentclass{article}

\usepackage{times}

\usepackage{graphicx} 
\usepackage{floatrow}
\usepackage{subfig}
\usepackage{courier}
\usepackage[font={small}]{caption}

\usepackage{tabu}
\usepackage{tabularx}
\usepackage[toc,page]{appendix}
\usepackage{xcolor, colortbl}
\usepackage[export]{adjustbox}

\usepackage[numbers]{natbib}

\usepackage{algorithm}
\usepackage{algorithmic}

\usepackage{hyperref}

\newcommand{\eb}{\boldsymbol{e}}

\newcommand{\vb}{\boldsymbol{v}}
\newcommand{\wb}{\boldsymbol{w}}
\newcommand{\xb}{\boldsymbol{x}}
\newcommand{\yb}{\boldsymbol{y}}

\newcommand\eqdef{\mathrel{\stackrel{\makebox[0pt]{\mbox{\normalfont\tiny def}}}{=}}}
\newcommand{\Mod}[1]{\ (\mathrm{mod}\ #1)}

\usepackage{amsmath}
\usepackage{amsthm}
\usepackage{amsfonts}
\usepackage{amssymb}

\usepackage[T1]{fontenc}
\usepackage{times}

\usepackage{geometry}
\usepackage[explicit]{titlesec}

\usepackage{tikz}
\usetikzlibrary{bayesnet,shapes,arrows}

\newcommand{\E}{\mathrm{E}}

\newcommand\tab[1][0.5cm]{\hspace*{#1}}

\title{
\hrule height 3pt
\vspace{6mm}
\textbf{Stochastic EM for Shuffled Linear Regression}
}

\author{
  \textbf{Abubakar Abid}\thanks{Department of Electrical Engineering, Stanford University. Correspondence to: James Zou <jamesz@stanford.edu>} \tab \textbf{James Zou\footnotemark[1]}
}

\date{\vspace{-14mm}}

\begin{document}

\maketitle
\kern0.5cm \hrule height 1pt \kern0.6cm

\begin{abstract} 
\normalsize
We consider the problem of inference in a linear regression model in which the relative ordering of the input features and output labels is not known. Such datasets naturally arise from experiments in which the samples are shuffled or permuted during the protocol. In this work, we propose a framework that treats the unknown permutation as a latent variable. We maximize the likelihood of observations using a stochastic expectation-maximization (EM) approach. We compare this to the dominant approach in the literature, which corresponds to hard EM in our framework. We show on synthetic data that the stochastic EM algorithm we develop has several advantages, including lower parameter error, less sensitivity to the choice of initialization, and significantly better performance on datasets that are only partially shuffled. We conclude by performing two experiments on real datasets that have been partially shuffled, in which we show that the stochastic EM algorithm can recover the weights with modest error.

\end{abstract} 

\section{Introduction}
\label{section:intro}
Linear regression is a widely used approach in statistics and machine learning to model the relationship between explanatory variables (\textit{input features}) and a continuous dependent variable (\textit{label}) \cite{stanton2001galton, seber2012linear}. In the classical setting, linear regression is applied to supervised datasets consisting of training examples that are fully and individually labeled. Many modern datasets do not meet this criteria, so in recent years, the question of inference from \textit{weakly-supervised} datasets has drawn attention in the machine learning community \cite{crandall2006weakly, li2013convex, zantedeschi2016beta}. In weakly-supervised datasets, data are neither entirely labeled nor entirely unlabeled: a subset of the data may be labeled, as is the case in semi-supervised learning, or the data may be implicitly labeled, as occurs, for example, in multi-instance learning \cite{zhou2011semi, zhou2007relation}. Weakly-supervised datasets naturally arise in situations where obtaining labels for individual data is expensive or difficult; often times, it is significantly easier to conduct experiments that provide partial information. 

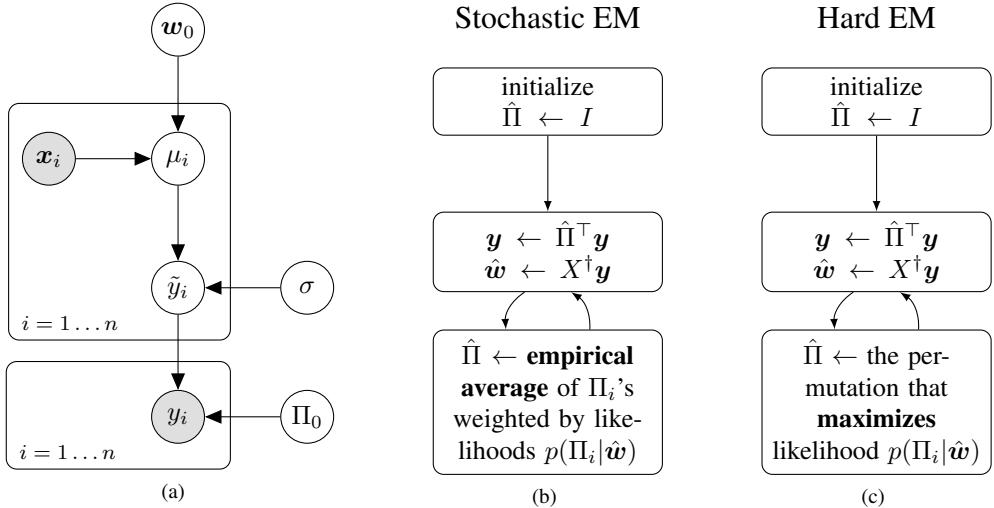
\begin{figure}[bt] 
\hfill
  \subfloat[\label{fig:1}]{%
      \centering
      \adjustbox{valign=c}{
      \tikz{ %
        \tikzset{plate caption/.append style={below right=0pt and 0pt of #1.south west}}
        \node[obs] (x) {$\xb_i$} ; %
        \node[latent, right=of x] (mu) {$\mu_i$} ; %
        \node[latent, above=of mu] (w) {$\wb_0$} ; %
        \node[latent, below=of mu] (yt) {$\tilde{y}_i$} ; %
        \node[latent, right=of yt] (sigma) {$\sigma$} ; %
        \node[obs, below=of yt] (y) {$y_i$} ; %
        \node[left=15mm of y] (inv) {\null} ; %
        \node[latent, right=of y] (pi) {$\Pi_0$}; %
        \plate[inner sep=0.25cm, xshift=0.08cm, yshift=0.12cm] {plate1} {(x) (yt)} {$i=1 \ldots n$}; %
        \plate[inner sep=0.25cm, xshift=0.08cm, yshift=0.12cm] {plate1} {(inv) (y)} {$i=1 \ldots n$}; %
        \edge {x} {mu} ; %
        \edge {w} {mu} ; %
        \edge {mu} {yt} ; %
        \edge {sigma} {yt} ; %
        \edge {yt} {y} ; %
        \edge {pi} {y} ; %
      }  } 
      }
    \hfill
  \subfloat[\label{fig:2}]{%


      \adjustbox{valign=c}{
        \tikzstyle{block} = [rectangle, draw, 
            text width=8em, text centered, rounded corners]
        \tikzstyle{line} = [draw, -late\xb']
        
        \begin{tikzpicture}[node distance = 2cm, auto]
            \node [block, label={[label distance=4mm]\large Stochastic EM}] (init) {initialize \\$\hat{\Pi} \leftarrow I$};
            \node [block, below of=init] (theta) {$\yb \leftarrow \hat{\Pi}^\top \yb$ $\hat{\wb} \leftarrow X^\dagger \yb$};
            \node [block, below of=theta] (pi) {$\hat{\Pi} \leftarrow$ \textbf{empirical average} of $\Pi_i$'s weighted by likelihoods $p(\Pi_i|\hat{\wb})$};
            \draw[-latex]  (init) edge (theta);
            \draw[-latex,bend right]  (theta) edge (pi);
            \draw[-latex,bend right]  (pi) edge (theta);
        \end{tikzpicture}
        }
    
  } 
    \hfill
  \subfloat[\label{fig:2}]{%

      \adjustbox{valign=c}{
        \tikzstyle{block} = [rectangle, draw, 
            text width=8em, text centered, rounded corners]
        \tikzstyle{line} = [draw, -late\xb']
        
        \begin{tikzpicture}[node distance = 2cm, auto]
            \node [block, label={[label distance=4mm]\large Hard EM}] (init) {initialize \\$\hat{\Pi} \leftarrow I$};
            \node [block, below of=init] (theta) {$\yb \leftarrow \hat{\Pi}^\top \yb$ $\hat{\wb} \leftarrow X^\dagger \yb$};
            \node [block, below of=theta] (pi) {$\hat{\Pi}$ $\leftarrow$ the permutation that \textbf{maximizes} likelihood $p(\Pi_i|\hat{\wb})$};
            \draw[-latex]  (init) edge (theta);
            \draw[-latex,bend right]  (theta) edge (pi);
            \draw[-latex,bend right]  (pi) edge (theta);
        \end{tikzpicture}
        }

  }    \hfill\null
  \caption{\textbf{Generative model and EM framework for shuffled linear regression} (a) we show the generative model for the data as a graphical model, where observed data are shaded and unobserved variables are clear. The parameters outside of the rounded rectangles are shared across all $n$ data points. The relationship between  variables is as follows: $\mu_i = \xb_i\cdot \wb_0$, $\tilde{y}_i = \mathcal{N}(\mu_i,\sigma^2)$, $\yb = \Pi_0 \cdot \tilde{\yb}$. (b). In the stochastic EM approach we propose, $\Pi_0$ is treated as a latent variable whose expectation is approximated at every step. Using the expected value of $\Pi_0$ and the pseudoinverse $X^{\dagger}$, the most likely value of $\wb_0$ is computed. (c). This is in contrast to the hard EM approach which uses an alternating maximization approach to estimate $\wb_0$ based on \textit{only} the most likely value of $\Pi_0$ and $X^{\dagger}$. \label{fig:model}}
\end{figure}

In this paper, we study one specific case of weak supervision: shuffled linear regression, in which all of the labels are observed, but the mutual ordering between the input features and the labels is unknown. Shuffled linear regression, then, can be described as a variant of traditional linear regression in which the labels are additionally perturbed by an unknown permutation. More concretely, the learning setting is defined as follows: we observe (or choose) a matrix of input features $X \in \mathbb{R}^{n \times d}$, and observe a vector of output labels $\yb \in \mathbb{R}^{n}$ that is generated as follows: 
\begin{equation}
\label{eqn_model}
\yb = \Pi_0 X \wb_0 + \eb
\end{equation}
where $\Pi_0$ is an unknown $n \times n$ permutation matrix, $\wb_0 \in \mathbb{R}^{d}$ are unknown coefficients, and $\eb \in \mathbb{R}^{n}$ is additive Gaussian noise drawn from $\mathcal{N}(0,\sigma^2)$. Here, $n$ is the number of data points, and $d$ is the dimensionality of the input features. The generative model described by (\ref{eqn_model}) and illustrated in Figure \ref{fig:model}(a) frequently results from experiments that simultaneously analyze a large number of objects. Consider an example.

\paragraph*{Example: Flow Cytometry.} The characterization of cells using flow cytometry usually proceeds by suspending cells in a fluid and flowing them through a laser that excites components within or outside the cell. By measuring the scattering of the light, various properties of the cells can be quantified, such as granularity or affinity to a particular target \cite{shapiro2005practical}. These properties (or labels) may be explained by features of the cell, such as its gene expression, a relationship that we are interested in inferring. However, because the \textit{order} of the cells as they pass through the laser is unknown, traditional inference techniques cannot be used to associate these labels with features of the cells that are measured separately from the flow cytometry.

\paragraph*{} Shuffled regression is also useful in other contexts where the order of measurements is unknown; for example, it arises in signaling with identical tokens \cite{rose2014signaling}, simultaneous pose-and-correspondence estimation in computer vision \cite{david2004softposit}, and in relative dating from archaeological samples \cite{robinson1951method}. An important setting where the feasibility of shuffled regression raises concern is data de-anonymization, such as of public medical records, which are sometimes shuffled to preserve privacy \cite{li2004protection}.

In this paper, we propose a latent variable framework to model the problem of shuffled regression. The framework naturally leads to an expectation-maximization (EM)-based algorithm to infer the weights. While this algorithm remains intractable to compute exactly, we develop a stochastic approximation, outlined in Fig. \ref{fig:model}(b), and described in Section \ref{section:stochastic-em}. Furthermore, we show that the dominant approach in literature, using alternating optimization to simultaneously estimate both $\Pi_0$ and $\wb_0$, corresponds to an alternative \textit{Hard EM} algorithm in this framework (see Fig. \ref{fig:model}(c) and Section \ref{section:hard-em}). 

In Section 3, we compare the performance of our algorithm, which we refer to as \textit{Stochastic EM}, to Hard EM on synthetic data, showing that Stochastic EM has a number of favorable properties, such as lower parameter error, more consistent results across permutations of the data, as well significantly improved performance with partially shuffled data. We then apply both algorithms to two real datasets that have been partially-shuffled in Section \ref{section:real}, finding that Stochastic EM produces lower MSE on test data.

\paragraph*{Related Work.} The problem of linear regression with shuffled labels has drawn significant interest in recent literature, where it has also been referred to as ``linear regression without correspondence'' \citep{hsu2017linear} or ``linear regression with permuted data'' \citep{pananjady2017denoising}. The most well-studied estimator finds the weights $\wb$ that produce the minimum squared distance between $X\cdot\wb$ and \textit{any} permutation of $\yb$ \citep{pananjady2017denoising,pananjady2016linear,wang2017maximum}. Specifically:

\begin{equation}
\hat{\wb} = \arg \min_{\wb} \min_{\Pi \in \mathcal{P}}  || \yb - \Pi X \wb ||^2, \label{eq:2}
\end{equation}
where $\mathcal{P}$ is the set of all $n \times n$ permutation matrices. While this estimator is, in general, NP-hard to compute \citep{pananjady2016linear}, optimization algorithms for (\ref{eq:2}) have been proposed based on alternating optimization of $\wb$ and $\Pi$ \citep{wang2017maximum,abid2017linear}. We will show that the alternating optimization approach is equivalent to performing Hard EM in our framework, and does not perform as well as Stochastic EM method.

There has also been prior related work on determining classifiers using proportional labels \cite{stolpe2011learning, yu2013propto}. In this setting, the authors aim to predict discrete classes that label instances belong to from training data that is provided in groups and only the proportion of each class label is known. This can be viewed as the corresponding problem of ``shuffled classification'' and efficient algorithms have been proposed for the task. However, these methods cannot be easily applied when labels are continuous, motivating the need for algorithms for shuffled regression.

\section{EM-Based Algorithms\label{section:em}}

Our algorithms for estimating $\wb$ are derived by maximizing the probability of observing $X$ and $\yb$, while treating the permutation matrix $\Pi_0$ as a latent variable. The maximum log-likelihood estimate of $\wb_0$ is:

\begin{equation}
\hat{\wb} = \arg\max_{\wb} \log p(X,\yb|\wb) = \arg\max_{\wb} \log \left( \sum_{\Pi}{p(X,\yb,\Pi|\wb)} \right) \label{eq:3}
\end{equation}

Directly maximizing the expression on the right hand side of (\ref{eq:3}) is difficult because of the summation inside the logarithm.
Following the standard expectation-maximization (EM) approach \citep{dempster1977maximum}, we rewrite the objective function on the right hand side of (\ref{eq:3}) in terms of a distribution $q(\Pi)$ over the permutations as
\begin{align}
    &\;\;\;\; \log \left( \sum_{\Pi}{p(X,\yb,\Pi|\wb)} \right)\nonumber \\
    &= \log \left( \sum_{\Pi}{ q(\Pi) \frac{p(X,\yb,\Pi|\wb)}{q(\Pi)}} \right)\nonumber \\ 
    &= \log \left( \E_q \left[ \frac{p(X,\yb,\Pi|\wb)}{q(\Pi)} 
    \right] \right) \nonumber\\
    \text{\footnotesize(using Jensen's inequality)}&\ge \E_q \left[ \log \left(\frac{p(X,\yb,\Pi|\wb)}{q(\Pi)} 
    \right) \right]  
\end{align}
We now have a lower bound on the log-likelihood function that we can optimize both over the parameters $\wb$ and the distribution $q(\cdot)$. We do this in an iterative process where we alternatively fix either $\wb$ (E-step) or $q(\cdot)$ (M-step), and optimize over the other. Let us denote by $\hat{\wb}_m$ and $\hat{q}_m$ the estimates on the $m^{\text{th}}$ iteration of EM. Then the corresponding updates are:
\begin{align}
    \textbf{E-step:}\;\;& \hat{q}_{m+1} \leftarrow p(\Pi|X,\yb,\hat{\wb}_m) \\
    \textbf{M-step:}\;\;& \hat{\wb}_{m+1} \leftarrow \arg \max_{\wb} \E_{\hat{q}_m} \left[ \log p(X,\yb,\Pi|\wb) \right] \label{eq:m-step}
\end{align}
Furthermore, it can be shown (see Appendix \ref{app:derivation-mstep}) that (\ref{eq:m-step}) is equivalent to:
\begin{align}
     \hat{\wb}_{m+1} \leftarrow (X^{\top}X)^{-1}X^\top \left( \E_{\hat{q}_m} \left[\Pi^\top \right] \yb \right) \label{eq:15}
\end{align}
Thus, the M-step reduces to performing an ordinary least-squares regression on a weighted average of permutations of $\yb$. However, note that the M-step cannot be performed exactly, since the expectation across the distribution of permutation matrices consists of $n!$ terms, making it prohibitively expensive to compute. The two variants of the EM algorithms that we discuss differ in the approximations used in place of this expectation.

\subsection{Stochastic EM}
\label{section:stochastic-em}

In Stochastic EM, we use the Monte Carlo technique to replace the expectation over the permutations by an empirical average\footnote{Note that the empirical average of permutation matrices $\hat{\Pi}_m^\top$ is not a permutation matrix itself}:
\begin{align}
    \hat{\Pi}^\top_m \eqdef \E_{\hat{q}_m} \left[\Pi^\top \right] \approx \sum_{a=1}^{S} \Pi^\top_a \label{eqn:13},
\end{align}
where the samples $\Pi^\top_a$ are drawn according to their likelihood probabilities $p(\Pi_a|X,\yb,\hat{\wb}_m)$. Such samples can be drawn efficiently using the Metropolis-Hastings sampling technique on a Markov Chain defined over the set of permutations $\mathcal{P}$. We define the Markov Chain to consist of transitions between pairs of permutations that differ in exactly 2 positions.  To use the Metropolis-Hastings algorithm, we compute the ratio of likelihoods of permutations $\Pi_a$ and $\Pi_b$:
\begin{align}
\alpha(a,b) \eqdef \frac{q(\Pi_a)}{q(\Pi_b)} &= \frac{p(\Pi_a|X,\yb,\hat{\wb}_m)}{p(\Pi_b|X,\yb,\hat{\wb}_m)}\nonumber \\
&= \frac{p(\Pi_a,X,\yb,\hat{\wb}_m)}{p(\Pi_b,X,\yb,\hat{\wb}_m)} \frac{\sum_{\Pi\in\mathcal{P}} p(\Pi,X,\yb,\hat{\wb}_m)}{\sum_{\Pi\in\mathcal{P}} p(\Pi,X,\yb,\hat{\wb}_m)} \nonumber \\ 
&= \frac{p(\Pi_a,X,\yb,\hat{\wb}_m)}{p(\Pi_b,X,\yb,\hat{\wb}_m)} \label{eq:19},
\end{align}
and make the transition from $\Pi_b$ to $\Pi_b$ with probability $\max(\alpha(a,b),1)$. Under the assumption that all permutations are equally likely \textit{a priori}, and in the case of Gaussian noise, (\ref{eq:19}) is computed as
\begin{align}
    \frac{p(\Pi_a,X,\yb,\hat{\wb}_m)}{p(\Pi_b,X,\yb,\hat{\wb}_m)} = \frac{ \prod_{i=1}^n \exp{(-||(\Pi_a X)_i \cdot \hat{\wb}_m - y_i||^2/\hat{\sigma}_m^2})}{\prod_{i=1}^n \exp{(-||(\Pi_b X)_i \cdot \hat{\wb}_m - y_i||^2/\hat{\sigma}_m^2})}, \label{eq:prob-ratio}
\end{align}
where $(\Pi X)_i$ is the $i^{\text{th}}$ row of $\Pi X$. This expression can be further simplified since the products in the numerator and denominator of (\ref{eq:prob-ratio}) differ in at most 2 terms. Furthermore, the transition probabilities can be modified naturally to incorporate priors or constraints in the case of partial shuffling of data. We show how to do this in the case of independent shuffling within groups of the data in Section \ref{section:real}.

Evaluating (\ref{eq:prob-ratio}) does require obtaining an estimate for $\sigma^2$ at each step, which we have denoted as $\hat{\sigma}_m^2$. This estimate can be computed along with the estimate of $\wb_0$ in the M-step using ordinary least squares. In fact, for convenience, let us define an OLS function as follows: 
\begin{align}
\text{OLS}(X,\yb) \to (\wb, \sigma^2), \; \text{where} \;\;
\wb \eqdef (X^\top X)^{-1}X^{\top} \yb, \; \text{and} \; 
\sigma^2 \eqdef \frac{1}{n-d}||X\cdot \wb - \yb||^2 \label{eq:ols}
\end{align}
Then, we can write the E and M steps for Stochastic EM in terms of the OLS function as follows:
\begin{align}
    &\textbf{E-step:} \; \text{Approximate} \; \hat{\Pi}_m^T \; \text{using Metropolis-Hastings with} \;  \hat{\wb}_m,\hat{\sigma}_m \\
    &\textbf{M-step:} \; \hat{\wb}_{m+1}, \hat{\sigma}_{m+1} = \text{OLS}(X, \hat{\Pi}_m^T \yb)
\end{align}

The pseudocode for Stochastic EM is shown in Algorithm \ref{alg1}. The algorithm includes a burn in period to ensure that the Markov chain has mixed before sampling from it, as well as a gap between samples to promote sample independence.

\subsection{Hard EM}
\label{section:hard-em}

In the Hard EM algorithm, we avoid computing the summation over the set of permutation matrices by replacing the expectation in (\ref{eq:15}) by the probability corresponding to only the most likely permutation. Formally, on the $m^\text{th}$ step, we compute 
\begin{align}
\hat{\Pi}_m^\top = \arg \max_\Pi q(\Pi^\top) = \arg \max_\Pi p(\Pi^\top|X,\yb,\hat{\wb}_m) \label{eq:24}
\end{align}
Identifying the most likely permutation $\hat{\Pi}^T_m$ in (\ref{eq:24}) is an efficient computational problem, because it is equivalent to finding the permutation that minimizes the value of $||\Pi X \hat{\wb}_m - \yb||^2$. It can be shown that this permutation is the one that re-orders the elements of $\yb$ according to the order of $X \cdot \hat{\wb}_m$ (see Appendix \ref{app:derivation-hard}). Thus, we can write the E-step and M-step for Hard EM as:
\begin{align}
    &\textbf{E-step:} \; \hat{\Pi}_m^T \; \text{is the permutation that re-orders} \; \yb \; \text{according to the order of} \; X \cdot \hat{\wb}_m \\
    &\textbf{M-step:} \; \hat{\wb}_{m+1} = (X^\top X)^{-1}X^\top (\hat{\Pi}_m^T \cdot \yb)
\end{align}
Thus, Hard EM reduces to a coordinate ascent on $\Pi$ and $\wb$ in alternate steps. The pseudocode for Hard EM is shown in Algorithm \ref{alg2}. In practice, we repeat the algorithm with multiple initializations of $\hat{\wb}$ and then choose the estimate that produces the highest likelihood. 

We also note that Hard EM is equivalent to performing coordinate \textit{descent} on $\Pi$ and $\wb$ to minimize the objective function widely studied in the shuffled regression literature (see Related Works in Section \ref{section:intro}):
\begin{align}
\arg \min_{\wb} \min_{\Pi \in \mathcal{P}}  || \yb - \Pi X \wb ||^2. \label{eq:optimization}
\end{align}
It is for this reason that we use Hard EM as a benchmark with which to compare Stochastic EM. 
\renewcommand*\footnoterule{}

\begin{figure}[tb!]
 \begin{minipage}[t]{2.9 in}
 \begin{algorithm}[H]
\caption{Stochastic EM} 
\label{alg1} 
\begin{algorithmic}[1]
\small
\INPUT $X$, $\yb$; number of iterations $k$, sampling steps $s$, burn steps $s'$, gap between samples $g$
\STATE initialize $\hat{\wb}, \hat{\sigma}^2 \leftarrow \text{OLS}(X,\yb)$ \textcolor{gray}{// see (\ref{eq:ols})} 
\STATE initialize $\hat{\Pi} \leftarrow 0$, $\Pi_a \leftarrow I$ (both matrices are of size $d \times d$, where $d$ is the dimensionality of $X$)
\FOR{i $\leftarrow$ 1 \textbf{to} $k$} 
\FOR{j $\leftarrow$ 1 \textbf{to} $s$} 
\STATE let $\Pi_{b}$ be the result of swapping two randomly-chosen rows of $\Pi_a$
\IF {$\frac{q(\Pi_b)}{q(\Pi_a)} \ge 1$ \textcolor{gray}{// see (\ref{eq:19})}
}
\STATE $\Pi_a \leftarrow \Pi_b$
\ELSE 
\STATE with probability $\frac{q(\Pi_b)}{q(\Pi_a)}$, $\Pi_a \leftarrow \Pi_b$
\ENDIF
\IF {$j > s'$ and $j \equiv 0 \Mod{g}$}
\STATE $\hat{\Pi} \leftarrow \hat{\Pi} + \Pi_a$ 
\ENDIF
\ENDFOR
\STATE $\hat{\Pi} \leftarrow \hat{\Pi}/s$; $\hat{\yb} \leftarrow \hat{\Pi}^\top\hat{\yb}$;  $\hat{\wb}, \hat{\sigma}^2  \leftarrow \text{OLS}(X,\yb)$
\ENDFOR
\STATE \bf{return} $\hat{\wb}$

\end{algorithmic}
\end{algorithm}
 \end{minipage}
 \hfill
  \begin{minipage}[t]{2.9in}
\begin{algorithm}[H]
\caption{Hard EM} 
\label{alg2} 
{\fontsize{10}{13}\selectfont
\begin{algorithmic}[1]
\small
\INPUT $X$, $\yb$, number of iterations $k$
\STATE initialize\footnote{initialization is typically performed by doing ordinary least-squares regression with $X$ and a random permutation for $y$. See Sections 3 and 4 for more experiment-specific details.} $\hat{\wb}$
\STATE sort $\yb$ in ascending order
\FOR{i $\leftarrow$ 1 \textbf{to} $k$} 
\STATE $\hat{\yb} \leftarrow X \cdot \hat{\wb}$
\STATE $\hat{\Pi} \leftarrow$ the permutation matrix that sorts $\hat{\yb}$ in ascending order
\STATE $X \leftarrow \hat{\Pi} \cdot X$ 
\STATE $\hat{\wb} \leftarrow (X^\top X)^{-1}X^{\top} \yb$ 
\ENDFOR
\STATE \bf{return} $\hat{w}$
\end{algorithmic}
}
\end{algorithm}
 \end{minipage}
 \hfill
\end{figure}

\section{Comparing Stochastic and Hard EM on Synthetic Data}

In this section, we compare the performance of Stochastic EM to Hard EM on synthetic data and find that the Stochastic EM algorithm compares favorably to Hard EM in three respects: (1) Stochastic EM produces estimates with lower parameter error on many datasets, (2) Stochastic EM produces results more consistently across permutations of the input data, and (3) Stochastic EM produces significantly better results when the data is only partially shuffled. We demonstrate these results in this section. 

In each experiment, we run Hard EM with $n$ different initializations. We set the number of iterations $(k)$ for both algorithm to be $50$. We set the number of sampling steps $(s)$ for Stochastic EM to be $n \log{n}$, the number of burn steps to be $n$, the gap between samples to be $n/10$. This roughly equalizes the run-time of the Hard and Stochastic EM algorithms to be $O(n^2 \log{n})$. Each synthetic dataset consists of an feature matrix $X$ whose entries are uniformly drawn from $\mathcal{N}(0,1)$, a weight vector $\wb_0$ whose entries are drawn from $\mathcal{N}(0,1)$. Parameter error is measured as $||\hat{\wb} - \wb_0||_2$.

\begin{figure}[bt] 
    \centering
    \subfloat[]{{\includegraphics[width=5cm]{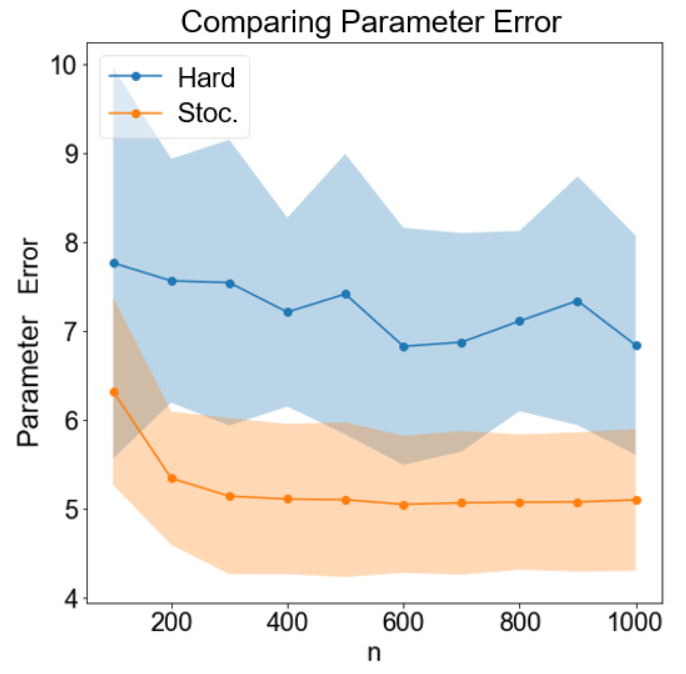} }}%
    \qquad
    \subfloat[]{{\includegraphics[width=5cm]{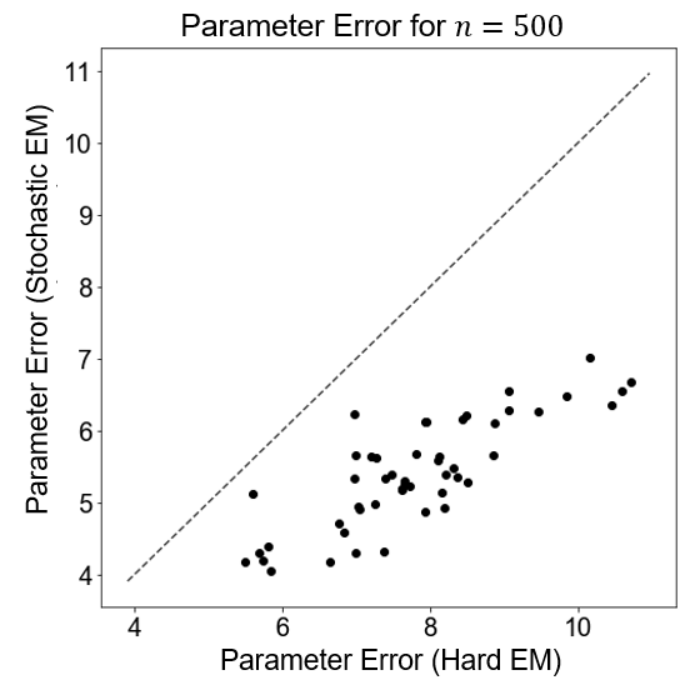} }}%
    \caption{\textbf{Parameter error for Hard EM and Stochastic EM}: (a) we generate random datasets of dimensionality $d=30, \sigma^2=1$ and different values of $n$ from 100 to 1,000. We apply both Hard EM and Stochastic EM and measure the parameter error ($||\hat{\wb} - \wb_0||_2$) across 10 trials. The graph shows the average and standard deviation parameter errors. (b) Here, we restrict ourselves to datasets of size $n=500, d=30$. We generate 50 synthetic datasets, in the same way as in (a). Each dot in the graph shows the parameter error produced by Hard EM and Stochastic EM for one dataset.}%
    \label{fig:parameter-errors}%
\end{figure}

\paragraph*{Magnitude of Parameter Error} One important advantage of the Stochastic EM is its ability to recover the original weights more accurately than Hard EM. We compared both algorithms systematically and found that across many synthetic datasets, Stochastic EM recovers the original weights more accurately than Hard EM. We show representative results in Fig. \ref{fig:parameter-errors}(a) for $d=30$ and different values of $n$. Furthermore, for the specific value of $n=500$, we randomly generate 50 synthetic datasets and show the parameter error for Hard and Stochastic EM. For all 50 datasets, the error for Stochastic EM is smaller than that of Hard EM (Fig. \ref{fig:parameter-errors}b). 
Additional results are included in Appendix \ref{appendix:results}.

\begin{figure}[!tb] 
    \centering
    \includegraphics[width=10cm]{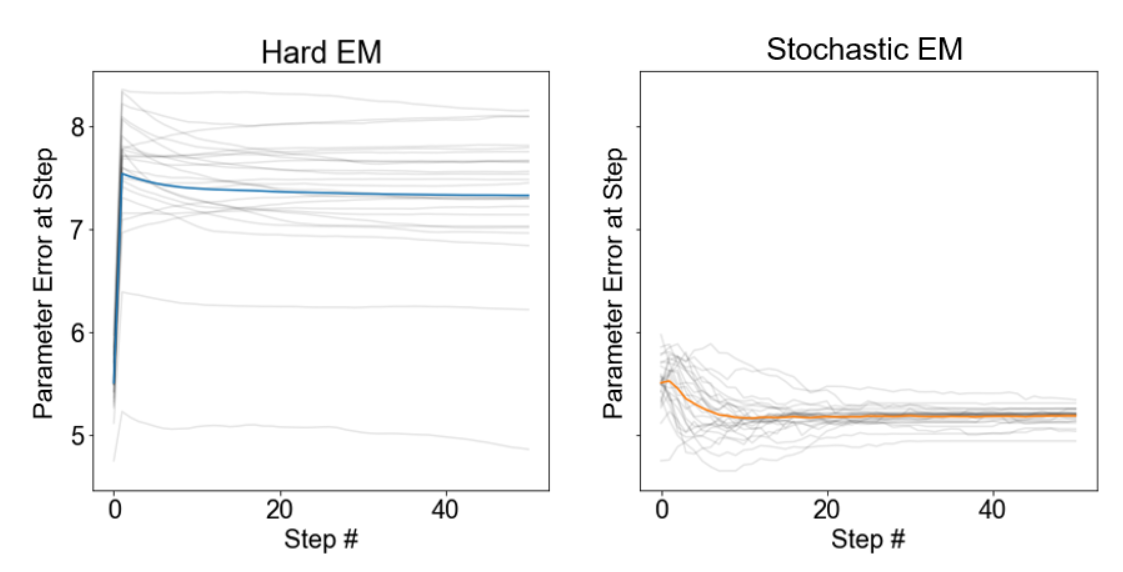}%
    \caption{\textbf{Consistency Across Permutations}: In this figure, we show the parameter errors of Hard EM (left) and Stochastic EM (right) at every step of the algorithm when performed on a synthetic dataset ($n=250, d=20$). Each gray line represents one run of the algorithm on a different permutation of the same dataset. The blue and orange lines represent the average parameter error across all 25 runs for Hard EM and Stochastic EM separately (the y-axes are shared). We find that the parameter error differs greatly among runs of Hard EM, and varies significantly less for each run of Stochastic EM.}%
    \label{fig:consistency}%
\end{figure}

\paragraph*{Consistency Across Permutations of Data} The initialization of $\wb$ for Stochastic EM is determined by the relative ordering of the input data $X$ and $\yb$. In contrast, Hard EM is repeated with many random initializations and so the final estimate would be expected to depend less on the initial ordering of the data. However, we actually find that Stochastic EM produces consistent weights across permutations of the labels, while Hard EM produces very different results, even when run with $n$ different initializations each time. Fig. \ref{fig:consistency} shows the parameter error of the weights at each step of Hard EM and Stochastic EM, when the algorithms are run on 25 different initial orderings of the labels.

\paragraph*{Performance on Partially Shuffled Data} In some cases, observed labels $\yb$ may be only a \textit{partially} shuffled version of the latent labels $\tilde{\yb}$. We simulate this scenario by starting with a vector of labels $\yb_0$ that is the same as $\tilde{\yb}$. We then randomly pick two elements in $\yb_0$ and swap them to generate $\yb_1$. We continue this process to generate progressively shuffled labels. We then feed each of these versions $\yb_i$ of the label into Hard EM\footnote{To make it a fair comparison, we ensure that in one of the $n$ runs of Hard EM, $\hat{\wb}$ is initialized to be the value of $(X^\top X)^{-1}X^\top \yb_i$, the same initialization as Stochastic EM.} and Stochastic EM and measure the parameter error. We find that Hard EM reports high error, even when the labels are only slightly shuffled, while Stochastic EM results in significantly lower error when the labels are not very shuffled.

\section{Experiments on Real Datasets}
\label{section:real}
In this section, we carry out experiments on public datasets that are collected in two settings where shuffling of data occurs: \textit{gated flow cytometry} and \textit{anonymization of sensitive information}. In each experiment, we compare Hard EM and Stochastic EM to two controls: a baseline that performs ordinary least-squares directly on the shuffled data, and performing ordinary least-squares on the \textit{unshuffled} data. The former serves as a negative control, while the latter (which can be only be performed if the true ordering is known, as is the case in these datasets) serves as a positive control to benchmark the maximum possible performance.

Furthermore, to make inference more tractable and control for overfitting (see discussion in Section \ref{section:discussion}), we do not entirely shuffle the datasets, but rather shuffle the data within a number, $G$, of groups, effectively creating $G$ subsets of features that are mapped to $G$ subsets of shuffled labels. As we explain below, this formulation is quite natural for both of the scenarios we consider. This additional structure allows us to improve both the Hard EM and Stochastic EM by only considering those permutations that map labels to the input features within the proper subset. The pseudocode for these modified algorithms are provided in Appendix \ref{appendix:enhanced} and it is these versions of algorithms that are used in this section.

\begin{figure}[!bt] 
\floatbox[{\capbeside\thisfloatsetup{capbesideposition={right,top},capbesidewidth=9cm}}]{figure}[\FBwidth]
{\caption{\textbf{Partially Shuffled Data}: In this figure, we show the parameter errors of Hard EM (blue) and Stochastic EM (orange) on a dataset ($n=200, d=20$) that is originally unshuffled, and is gradually shuffled by iteratively swapping two elements of $\yb$ (we denote the number of iterations on the x-axis). We compute the parameter error after every five swaps. The graph shows the mean (dots) and standard deviation (shaded region), across five series of shuffles, of the parameter errors as a function of the number of shuffles. We find that Hard EM performs very poorly even when the dataset is only slightly shuffled, unlike Stochastic EM, which reports significantly improved performance when the labels are only slightly shuffled.}\label{fig:partially}}
{\includegraphics[width=4.5cm]{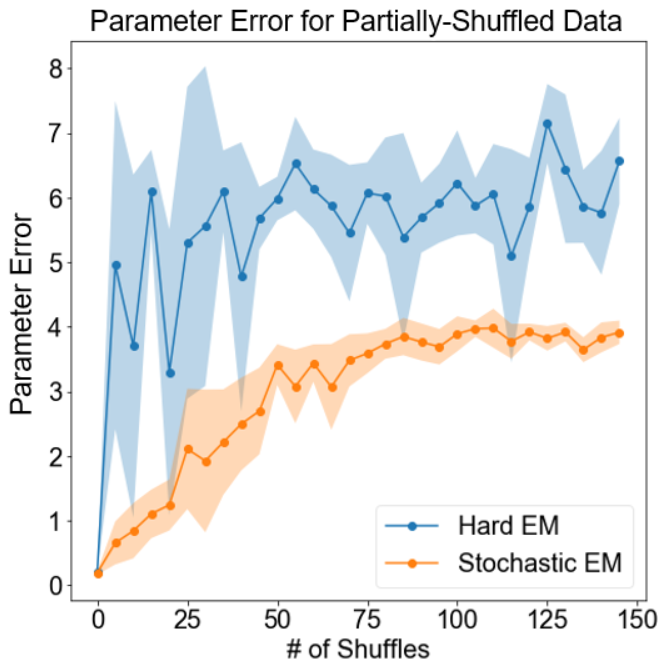}}
\end{figure}

\subsection{Gated Flow Cytometry for Aptamer Affinity}
\label{subsection:aptamer}

\paragraph*{Background.} Flow cytometry was introduced in Section \ref{section:intro}. In this subsection, we consider a modification referred to as gated flow cytometry: First, particles are suspended in fluid and flowed through a cytometer; then, the cytometer analyzes the properties of each particle, and uses ``gates'' to sort the particles into one of many bins. This sorting provides \textit{partial} ordering information, as it provides the analyst a list of labels for all of the particles in each bin. With enough bins, it would be possible to determine the ordering information completely, but generally, it is only practical to set up 3-4 bins.

Gated flow cytometry has many use cases, but the particular application we consider is measurement of aptamer affinity. Aptamers are short DNA sequences that bind to target molecules, and it is of interest to characterize affinity of aptamers to a particular target and to determine the dependence of affinity on the nucleotide motifs (e.g. ``AGG'' or ``CC'') present in aptamers. This analysis is useful in designing aptamers to have particularly high or low affinities to the target. Gated flow cytometry can be used to measure aptamer affinity by suspending a pool of aptamers in a solution of fluorescently-labeled targets. By measuring the fluorescence of the particles as they flow through the cytometer, a histogram of affinities is produced. By taking the aptamers (that have been sorted into bins) and then sequencing them, the corresponding features can be recovered.

\begin{figure}[bt] 
    \centering
    \subfloat[]{{\includegraphics[width=5cm]{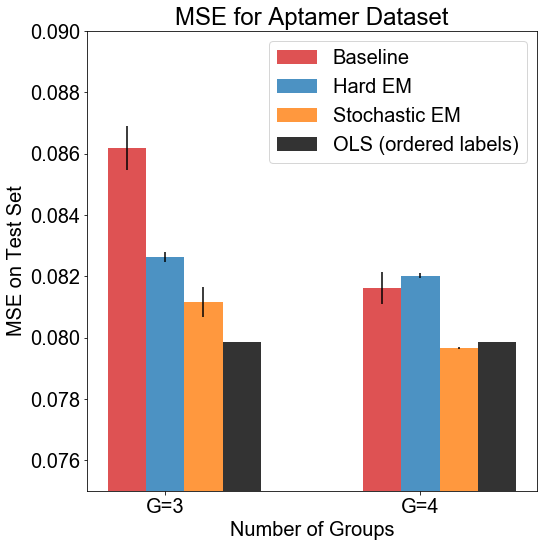} }}%
    \qquad
    \subfloat[]{{\includegraphics[width=5cm]{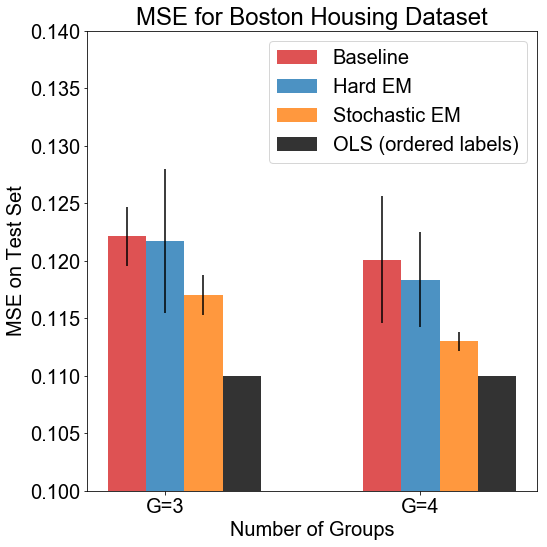} }}%
    \caption{\textbf{Experiments on real world datasets}: (a) We applied Stochastic EM and Hard EM on the aptamer datasets described in Section \ref{subsection:aptamer}, with 3 and 4 groups. Stochastic EM performs better than both the random baseline and Hard EM. In fact with, $G=4$ groups, Stochastic EM performs as well as OLS with known ordering. (b) Here, we apply both algorithms to the Boston housing dataset. Stochastic EM still outperforms the baseline and Hard EM on average, but the results are more variable. Each experiment is repeated five times, and standard deviation error bars are shown.}%
    \label{fig:real-world}%
\end{figure}

\paragraph*{Analysis.} The dataset we used is a publicly available aptamer affinity dataset \cite{knight2009array} with $n=5,000$ sequences, each labeled with its affinity to a target. We preprocessed the sequences to $d=84$ dimensions by counting the number of nucleotide subsequences of length 1, 2, and 3 in each sequence. This is a common method to featurize aptamers. The labels were normalized to range from 0 to 1. Then, the data were sorted in order according to their labels and separated into $G = 3$ or $4$ equal-sized groups, so that on average, each group consisted of labels that spanned a range of size $1/G$. The labels were then randomly shuffled within each group. To simulate sorting noise, 1\% of the data points were shuffled between bins. We then divided into training and test sets (80-20 split). Both Hard and Stochastic EM were applied to the input features and shuffled labels to obtain regression coefficients from the training set. These regression coefficients were used to make affinity predictions for each input sequence in the test set and the mean-squared error (MSE) between the predicted labels and actual labels was recorded (see Fig. \ref{fig:real-world}a).


\subsection{Partially Anonymized Housing Prices}
\label{subsection:housing}

\paragraph*{Background.} Another application of shuffled linear regression is in the analysis of datasets that contain labels that are permuted to preserve anonymity \citep{li2004protection}. The permutation makes inference difficult as the correspondence between features and labels is unknown. In this example, we analyze whether the shuffled regression framework can be applied to infer the relationship between features and a sensitive label. In our example, the sensitive label is socioeconomic status of the population of a town (specifically, the percent of low-income residents), and the features are other public characteristics about the town. We we divided the data according to one of the features, ``median housing price'' to create $G = 3$ or $4$ \textit{zones}. We model the scenario that people may not be willing to provide socioeconomic information on a per-town level, but may agree to do so if the information is anonymized within each zone. We thus anonymized the labels within each \textit{data subset}, which provided us partial ordering information\footnote{Note that in the previous experiment (Section \ref{subsection:aptamer}), we shuffled the data within bins that separated the data by the values of the \textit{label}, while here, we shuffle the data within bins separated by the value of one of the \textit{features}.}.  

\paragraph*{Analysis.} The dataset that we used is the publicly available Boston Housing dataset ($n=506, d=13$) \citep{harrison1978hedonic}. We normalized the labels to range from 0 to 1. Then, the data were sorted in order according to the median housing price and separated into $G$ equal-sized groups, so that each group represented the anonymized housing. The labels were then randomly shuffled within each group. We then divided into training and test sets (80-20 split). Both Hard and Stochastic EM were applied to the input features and shuffled labels to obtain regression coefficients from the training set. These regression coefficients were used to make predictions for each data point in the test set and mean-squared error between the predicted labels and actual labels was recorded.

\section{Discussion}
\label{section:discussion}
Machine learning with weak supervision is a challenge that is increasingly common and important. We have proposed an algorithm based on a stochastic approximation to EM that improves statistical estimation in applications where we have incomplete correspondence between the input features and output labels. Our results show that this algorithm is more accurate and robust compared to the commonly-studied alternating optimization algorithm (equivalent to Hard EM) on both synthetic datasets as well real datasets with partial ordering information.

The reason that the Hard EM algorithm does not work as well as Stochastic EM is because the Hard EM algorithm only considers the \textit{most} likely permutation for a given value of $\hat{\wb}$. Therefore, it is extremely sensitive to initialization of $\hat{\wb}$. This is problematic because the optimization function that Hard EM minimizes, stated in (\ref{eq:optimization}), is highly non-convex (in fact, it is the minimum of $n!$ convex functions). As we show in Fig. \ref{fig:consistency}, this leads the Hard EM algorithm to find local minima that provide widely differing parameter errors, depending on the initialization of the algorithm. Even repeating the Hard EM with many restarts does not mitigate this problem for any reasonably large values for $n$ and $d$, and the average error remains much higher than for Stochastic EM, as shown in Fig. \ref{fig:parameter-errors}. For very small dataset sizes, where range of initializations of $\hat{\wb}$ is smaller, it may be possible to run the Hard EM with enough different initializations for it to converge to the (near-)globally optimal parameters. We show this in Appendix \ref{appendix:results}, for datasets of size $d=2$ and $d=5$. However, this approach is not scalable beyond the smallest of datasets.

A separate challenge that we face in shuffled linear regression is the challenge of \textit{overfitting}. In particular, noise in the data may cause a permutation other than $\Pi_0$ to produce a higher log-likelihood of observing the data than $\Pi_0$. This in turn may lead to an estimate of $\hat{\wb}$ that is far from $\wb_0$. This is particularly important when we perform inference on real datasets where the magnitude of noise is significant, or the noise is heteroscedastic, or the underlying relationship between $X$ and $\yb$ is not entirely linear. In Section \ref{section:real} of this work, we have proposed one possible solution to this problem: designing the generative process to produce \textit{multiple} data subsets that are independently shuffled. This additional structure reduces the potential to overfit as a given $\hat{\wb}$ must produce a high likelihood of observing \textit{all} of the data subsets. Another method to control over-fitting may be to hold out one or several of the data subsets (similar to a validation dataset) and use them to determine when to stop iterating Stochastic or Hard EM. These methods will be explored in future work. 

Despite improved performance of our approach relative to the alternating-optimization approach that is standard in the literature, we believe that an open area of research remains to develop algorithms that can be practically applied to real datasets that are completely shuffled. Although this remains challenging due to highly non-convex nature of the problem and potential to overfit discussed earlier, the EM-based framework we have developed may provide some guidance on how to produce more general algorithms for this problem.

\bibliographystyle{unsrt}
\bibliography{references}

\newpage
\begin{appendices}
\section{Additional Derivations}
\label{app:derivations}
\subsection{Derivation of the General M-step}
In Section \ref{section:em}, we claim that the M-step can be reduced from performing an expectation over $q_m(\Pi)$ to an expectation over $\Pi^\top$. The derivation is as follows:
\begin{align*}
     &\;\;\;\; \arg \max_{\wb} E_{\hat{q}_m} \left[ \log p(X,\yb,\Pi|\wb) \right]\\
     & = \arg \max_{\wb} \sum_{\Pi \in \mathcal{P}}\hat{q}_m(\Pi)\log p(X,\yb,\Pi|\wb)\\
     & = \arg \min_{\wb} \sum_{\Pi \in \mathcal{P}}\hat{q}_m(\Pi) || \Pi X \wb - \yb ||^2\\
     & = \arg \min_{\wb} \sum_{\Pi \in \mathcal{P}}\hat{q}_m(\Pi) (\wb^\top X^\top  X \wb - 2\yb^\top \Pi X \wb +  \yb \yb^\top) \\
     & = \arg \min_{\wb} (\wb^\top X^\top  X \wb - 2\yb^\top \left(\sum_{\Pi \in \mathcal{P}}\hat{q}_m(\Pi)\Pi \right) X \wb +  \yb \yb^\top)\\
     & = (X^{\top}X)^{-1}X^\top \left(\sum_{\Pi \in \mathcal{P}} \hat{q}_m(\Pi)\Pi^T \right) \yb\\ 
     & = (X^{\top}X)^{-1}X^\top \left( E_{\hat{q}_m} \left[\Pi^\top \right] \yb \right) \label{eq:15}
\end{align*}

\label{app:derivation-mstep}

\subsection{Derivation of the E-step for Hard EM}
\label{app:derivation-hard}

In Section \ref{section:hard-em}, we claim that the the permutation that minimizes the value of $||\Pi X \hat{\wb}_m - \yb||^2$ is the the permutation that reorders $\yb$ according to the order of $X \cdot \hat{\wb}_m$. The proof is below. 

Without loss of generality, assume that the elements of $\yb$ are sorted in ascending order, and for convenience define $\xb \eqdef X \cdot \hat{\wb}_m$. We proceed by contradiction, Let us assume that $|| \yb - \Pi \xb ||^2$ is minimized for a permutation of $\xb$, call it $\xb'$, that does not have its entries in ascending order. Then, there must be $i, j$ such that $i < j$ but $\xb'_i > \xb'_j$ where we use the notation $\vb_i$ to refer to the $i^{th}$ element of $\vb$. 
But then:

\begin{align*}
|| \yb - \xb' ||^2 &= \sum_k^n (\yb_k - \xb'_k )^2  \\
&= (\yb_{i} - \xb'_i)^2 + (\yb_{j} - \xb'_j)^2 +  \sum_{k \ne i,j}^n (\yb_{k } - \xb'_k )^2 \\
&= (\yb_{i} - \xb'_i)^2 + (\yb_{j} - \xb'_j )^2 +  \sum_{k \ne i,j}^n (\yb_{k } - \xb'_k )^2  \\
&\ge (\yb_{j} - \xb'_i)^2 + (\yb_{i} - \xb'_j )^2 +  \sum_{k \ne i,j}^n (\yb_{k } - \xb'_k )^2
\end{align*}
But this is a contradiction, because we have showed that we can achieve a smaller value by switching the $i^{th}$ and $j^{th}$ entry in $\xb'$.
\textit{Note}: The inequality in the last step follows from the fact that the difference between the final expression and the one before it can be written as:
$$ -2(\xb'_i \yb_i + \xb'_j \yb_j) + 2(\xb'_i \yb_j + \xb'_j \yb_i) = 2 (\yb_j - \yb_i) (\xb'_i - \xb'_j)  \ge 0, $$
where the last equality follows from the fact that  $(\yb_j - \yb_i) \ge 0$ because $\yb$ is sorted in ascending order and $(\xb'_i - \xb'_j) \ge 0$ by assumption.
$\,$

\section{Additional Results}
\label{appendix:results}

\begin{figure}[bt] 
    \centering
    \subfloat[]{{\includegraphics[width=5cm]{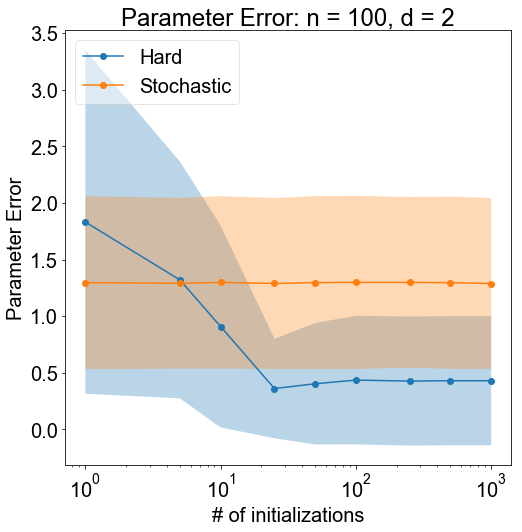} }}
    \qquad
    \subfloat[]{{\includegraphics[width=5cm]{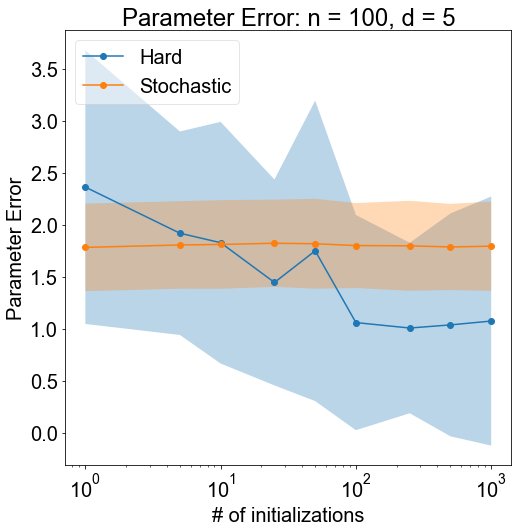} }} \\
    \subfloat[]{{\includegraphics[width=5cm]{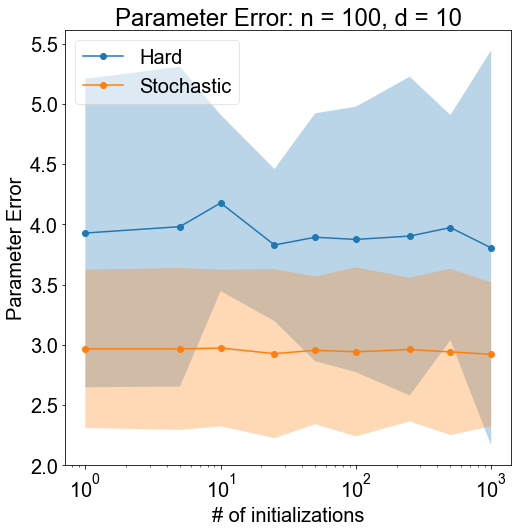} }}
    \qquad
    \subfloat[]{{\includegraphics[width=5cm]{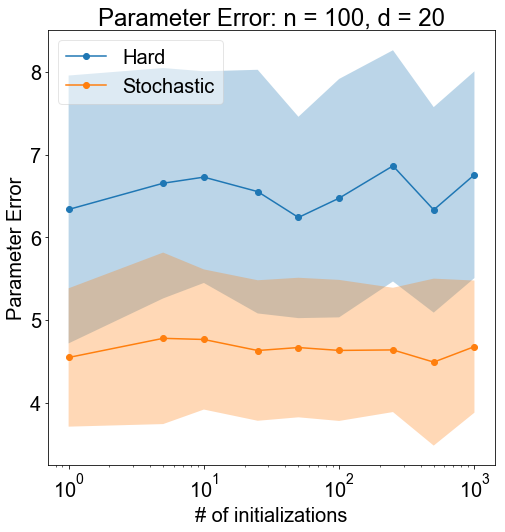} }}
    \caption{\textbf{Experiments with varying number of initializations}: In this series of experiments, we varied the number of initializations of Hard EM from 1 to 1000 ($n$ was fixed to 100). (a) For $d=2$, we observed markedly improved performance as the number of initializations increased. Eventually, Hard EM overtook Stochastic EM in performance. (b) A similar trend was observed for $d=5$. (c) For $d=10$, the performance of Hard EM did not improve, even the when number of initializations reached $10n$. (d) The same held true for $d=20$.}%
    \label{fig:inits}%
\end{figure}

In this section, we explore the effect of the number of initializations of Hard EM on its performance. We show that for datasets with small dimensionality, increasing the number of initializations of Hard EM can significantly increase performance, as shown in Fig. \ref{fig:inits}(a) and (b). This is because the with increased initiailzations, Hard EM is able to discover the globally optimally solution (or a solution that is nearly the global optimal) of the highly non-convex optimization problem stated in (\ref{eq:optimization}). However, this method does not scale beyond the smallest dataset sizes, as shown in Fig. \ref{fig:inits}(c) and (d). 

As before, each dataset consists of an feature matrix $X$ whose entries are uniformly drawn from $\mathcal{N}(0,1)$, a weight vector $\wb_0$ whose entries are drawn from $\mathcal{N}(0,1)$, and $\sigma = 0.3$. Parameter error is measured as $||\hat{\wb} - \wb_0||_2$.

\section{Psuedocode for Modified Algorithms}
\label{appendix:enhanced}

In Section \ref{section:real}, we modified Stochastic and Hard EM to take advantage of the fact that we were working with datasets that were shuffled within a number of groups, effectively creating several subsets of features that are mapped to subsets of shuffled labels. Let us denote the number of groups as $G$, the subsets of features $\{X^{(g)}\}_{g=1}^G$, and the subsets of labels $\{\yb^{(g)}\}_{g=1}^G$ so that the features $X^{(g)}$ correspond to the shuffled labels $\yb^{(g)}$. Then, the pseudocode for the modified Stochastic EM and Hard EM can be found in Algorithms \ref{alg3} and \ref{alg4} respectively.

\begin{figure}[tb!]
 \begin{minipage}[t]{2.9 in}
 \begin{algorithm}[H]
\caption{Modified Stochastic EM} 
\label{alg3} 
\begin{algorithmic}[1]
\small
\INPUT $\{ X^{(g)}, \yb^{(g)} \}_{g=1}^{G}$; \# of iterations $k$, sampling steps $s$, burn steps $s'$, gap between samples $h$
\STATE define $X$ to be the matrix formed by vertically concatenating $\{X^{(g)}\}_{g=1}^{G}$, and define $\yb$ similarly
\STATE initialize $\hat{\wb}, \hat{\sigma}^2  \leftarrow \text{OLS}(X,\yb)$
\STATE initialize $\{\hat{\Pi}^{(g)} \}_{g=1}^G \leftarrow 0$, $\{\Pi_a^{(g)} \}_{g=1}^G \leftarrow I$
\FOR{i $\leftarrow$ 1 \textbf{to} $k$} 
\FOR{j $\leftarrow$ 1 \textbf{to} $s$} 
\STATE Randomly select a $\Pi^{(g)}_a \in \{\Pi^{(g)}_a \}_{g=1}^G$
\STATE let $\Pi_{b}^{(g)}$ be the result of swapping two randomly-chosen rows of $\Pi_a^{(g)}$
\IF {$q(\Pi_b^{(g)})/q(\Pi_a^{(g)}) \ge 1$ }
\STATE $\Pi_a^{(g)} \leftarrow \Pi_b^{(g)}$
\ELSE 
\STATE with probability $\frac{q(\Pi_b^{(g)})}{q(\Pi_a^{(g)})}$, $\Pi_a^{(g)} \leftarrow \Pi_b^{(g)}$
\ENDIF
\IF {$j > s'$ and $j \equiv 0 \Mod{h}$}
\STATE $\hat{\Pi}^{(g)} \leftarrow \hat{\Pi}^{(g)} + \Pi_a^{(g)}$ 
\ENDIF
\ENDFOR
\STATE define $\hat{\Pi}$ to be the direct sum $\hat{\Pi}^{(1)} \bigoplus \ldots \bigoplus \hat{\Pi}^{(G)}$
\STATE $\hat{\Pi} \leftarrow \hat{\Pi}/s$; $\hat{\yb} \leftarrow \hat{\Pi}^\top\hat{\yb}$;  $\hat{\wb}, \hat{\sigma}^2  \leftarrow \text{OLS}(X,\yb)$
\ENDFOR
\STATE \bf{return} $\hat{\wb}$

\end{algorithmic}
\end{algorithm}
 \end{minipage}
 \hfill
\begin{minipage}[t]{2.9in}
\begin{algorithm}[H]
\caption{Modified Hard EM} 
\label{alg4} 
{\fontsize{10}{13}\selectfont
\begin{algorithmic}[1]
\small
\INPUT $\{ X^{(g)}, \yb^{(g)} \}_{g=1}^{G}$, number of iterations $k$
\STATE define $X$ to be the matrix formed by vertically concatenating $\{X^{(g)}\}_{g=1}^{G}$, and define $\yb$ similarly
\STATE initialize $\hat{\wb}$
\STATE sort $\yb$ in ascending order
\FOR{i $\leftarrow$ 1 \textbf{to} $k$} 
\FOR{g $\leftarrow$ 1 \textbf{to} $G$}
\STATE $\hat{\yb}^{(g)} \leftarrow X^{(g)} \cdot \hat{\wb}$
\STATE $\hat{\Pi}^{(g)} \leftarrow$ the permutation matrix that sorts $\hat{\yb}^{(g)}$ in ascending order
\ENDFOR
\STATE define $\hat{\Pi}$ to be the direct sum $\hat{\Pi}^{(1)} \bigoplus \ldots \bigoplus \hat{\Pi}^{(G)}$
\STATE $X \leftarrow \hat{\Pi} \cdot X$ 
\STATE $\hat{\wb} \leftarrow (X^\top X)^{-1}X^{\top} \yb$ 
\ENDFOR
\STATE \bf{return} $\hat{w}$
\end{algorithmic}
}
\end{algorithm}
 \end{minipage}
 \hfill
\end{figure}

\end{appendices}

\end{document}